\documentclass[letterpaper]{article} 
\usepackage{aaai23}  
\usepackage{times}  
\usepackage{helvet}  
\usepackage{courier}  
\usepackage[hyphens]{url}  
\usepackage[pdftex]{graphicx}
\usepackage{natbib}  
\usepackage{caption} 
\usepackage{algorithm}
\usepackage{algorithmic}
\usepackage{newfloat}
\usepackage{listings}
\DeclareCaptionStyle{ruled}{labelfont=normalfont,labelsep=colon,strut=off} 
\floatstyle{ruled}
\newfloat{listing}{tb}{lst}{}
\floatname{listing}{Listing}
 \nocopyright

\title{Cognitive Architecture Toward Common Ground Sharing Among Humans and Generative AIs: Trial on Model-Model Interactions in Tangram Naming Task}
\author{
    Junya Morita,\textsuperscript{\rm 1} Tatsuya Yui,\textsuperscript{\rm 1} Takeru Amaya,\textsuperscript{\rm 1} \\Ryuichiro Higashinaka\textsuperscript{\rm 2} Yugo Takeuchi\textsuperscript{\rm 1}
\\
}
\affiliations{
    \textsuperscript{\rm 1}Shizuoka University, 3-5-1, Johoku, Naka-ku, Hamamatsu 432-8011, Japan\\
        \textsuperscript{\rm 1}Nippon Telegraph and Telephone Corporation, 1-1 Hikarinooka,Yokosuka 239-0847, Japan\\

    j-morita@inf.shizuoka.ac.jp,
    yui.tatsuya.20@shizuoka.ac.jp,	amaya.takeru.19@shizuoka.ac.jp,\\
    ryuichiro.higashinaka@ntt.com,		takeuchi@inf.shizuoka.ac.jp
    
}

\usepackage{bibentry}

\begin{document}

\maketitle

\begin{abstract}
For generative AIs to be trustworthy, establishing transparent common grounding with humans is essential. 
As a preparation toward human-model common grounding, this study examines the process of model-model common grounding.
In this context, common ground is defined as a cognitive framework shared among agents in communication, enabling the connection of symbols exchanged between agents to the meanings inherent in each agent. This connection is facilitated by a shared cognitive framework among the agents involved.
In this research, we focus on the tangram naming task (TNT) as a testbed to examine the common-ground-building process. Unlike previous models designed for this task, our approach employs generative AIs to visualize the internal processes of the model. In this task, the sender constructs a metaphorical image of an abstract figure within the model and generates a detailed description based on this image. The receiver interprets the generated description from the partner by constructing another image and reconstructing the original abstract figure.
Preliminary results from the study show an improvement in task performance beyond the chance level, indicating the effect of the common cognitive framework implemented in the models. Additionally, we observed that incremental backpropagations leveraging successful communication cases for a component of the model led to a statistically significant increase in performance. These results provide valuable insights into the mechanisms of common grounding made by generative AIs, improving human communication with the evolving intelligent machines in our future society.
\end{abstract}

\section{Introduction}

Social applications of generative models\footnote{Following the symposium title, the 2023 {AAAI} Fall Symposium on Integrating Cognitive Architectures and Generative Models, we use the term ``generative model'' to refer to the computational technology that generates media naturally used by humans.} are growing these days. For example, web services such as ChatGPT \cite{openai:chatgpt} and Stable Diffusion \cite{rombach2021highresolution} summarize vast amounts of online data through lexical and graphical media that humans naturally handle in everyday life. The requests from the user are passed to the model as linguistically described prompts, and the model attempts to summarize the information according to the user's intention. The user then leverages the results as material for their original task and allocates cognitive resources to more creative endeavors.

However, a transparency problem has been pointed out regarding the situation surrounding these generative models. The construction of generative models requires huge data sets and high computational power, and it is currently accomplished by only a few large organizations, called Big Tech. Although explanations regarding the construction process are provided by them, the specific algorithms, parameters, and data selection methods remain implicit. This lack of transparency has led to a growing distrust regarding generative artificial intelligence (AI), particularly in EU \cite{madiega2021artificial}. This distrust of AI may result in a serious division of attitude around this technology in our society and narrow the possibilities of a future guided by this technology.

To overcome the abovementioned situation, a trustworthy framework connecting generative models and humans is required. The authors consider that such a framework should be built on the basis of collective public activities as represented by academic research. In other words, a framework to control generative models cannot be constructed by ignoring research related to {\it cognitive architectures}, which is a traditional academic field describing computational structures that are consistent with human intelligence \cite{Kotseruba2020}. The knowledge obtained in this field have accumulated in the form of academic papers and open-source software accessible to anyone in society and have been integrated in recent years in the form of common cognitive model \cite{laird2017standard}.

On the basis of the above idea, this paper presents a research plan for constructing a cognitive architecture that controls multiple generative models. Our long-term goal is to build a common ground shared between humans and models. As a test bed for the target situation, this research uses a task of common-ground building, which has been long studied in the field of cognitive science. Our aim here is to clarify the requirements of the architecture and the role of generative models in simulating the human common-ground building process. The following section presents the concrete task, our proposed architecture, and the model. We believe that such a basic academic endeavor will contribute to improving human communication with the evolving intelligent machines in our future society.

\section{Common-ground-building task}
\label{sec:contents-format1}

The need for a common ground in communication has been mentioned by various researchers \cite{clark1986referring,traum1994computational}. In a communication scenario, a sender's meaning conveyed by symbols is restored by the knowledge possessed by the receiver. Therefore, if there is no common cognitive framework\footnote{There are several terms to describe cognitive frameworks, such as schema, mindset, and reference frame.} between the sender and receiver, the communication cost is enormous. The existence of a common cognitive framework narrows down the meanings of polysemous symbols and enables quick communication through simplified expressions. Thus, this paper treats a common ground as a shared cognitive framework for attending features in encountered situation and coding them into linguistic expressions.

As a task to examine the process of common-ground building, a communication task using abstract figures called {\it tangrams} has been freaquently leveraged in the field of cognitive science \cite{clark1986referring}\footnote{Based on Clark's task using tangram, several researchers have invented tasks for revealing aspects of communicative process in a simplified situation. Such a movement of experimental studies is named as ``experimental semiotics \cite{galantucci2009experimental}.'' For example, the task inventing by Garrod et al is related to Clark's task by dealing with an emergence of graphical symbols from repetetive verbal communication \cite{garrod2007foundations}, but focusing on more advanced communication process.}. Tangrams are constructed by combining several simple figures. By considering tangrams as silhouettes of objects, various interpretations can be generated. The generation of these interpretations depends on the cognitive framework possessed by the perceiver at a given time. Therefore, a shared common ground is needed for the receiver to identify the tangram indicated by the sender.

Among several studies dealing with the task, we focus on the dialogue and process presented in Sudo et al.'s experiment \cite{sudo-etal-2022-speculative}, where pairs of participants aimed to agree on the naming of six tangrams. Hereafter, this experimental task will be referred to as the tangram naming task (TNT). Figure 1 shows an example of the tangram sets observed by the participants in the task. Although both participants are presented with the same set of tangrams, the placement and angles of the tangrams are different. In the task, participants cannot see each other's screens and are required to perform the task using only linguistic communication.

Sudo et al. analyzed utterances in TNT in terms of {\it holistic} and {\it analytic} processes. Examples of dialogue sequences are given in Table 1. Holistic utterances were those in which the shape of the tangram was metaphorically compared to a concrete object (e.g., {\it like Hokkaido}, {\it ball kicking}), and analytic utterances were those in which the tangram was referred to as a geometric figure (e.g., {\it a square and a triangle on each side}). Sudo et al.'s data showed that holistic utterances outnumbered analytic utterances throughout the experiment, and the difference widened as the session progressed.

\begin{figure}[t]
\centering
\includegraphics[width=1\columnwidth]{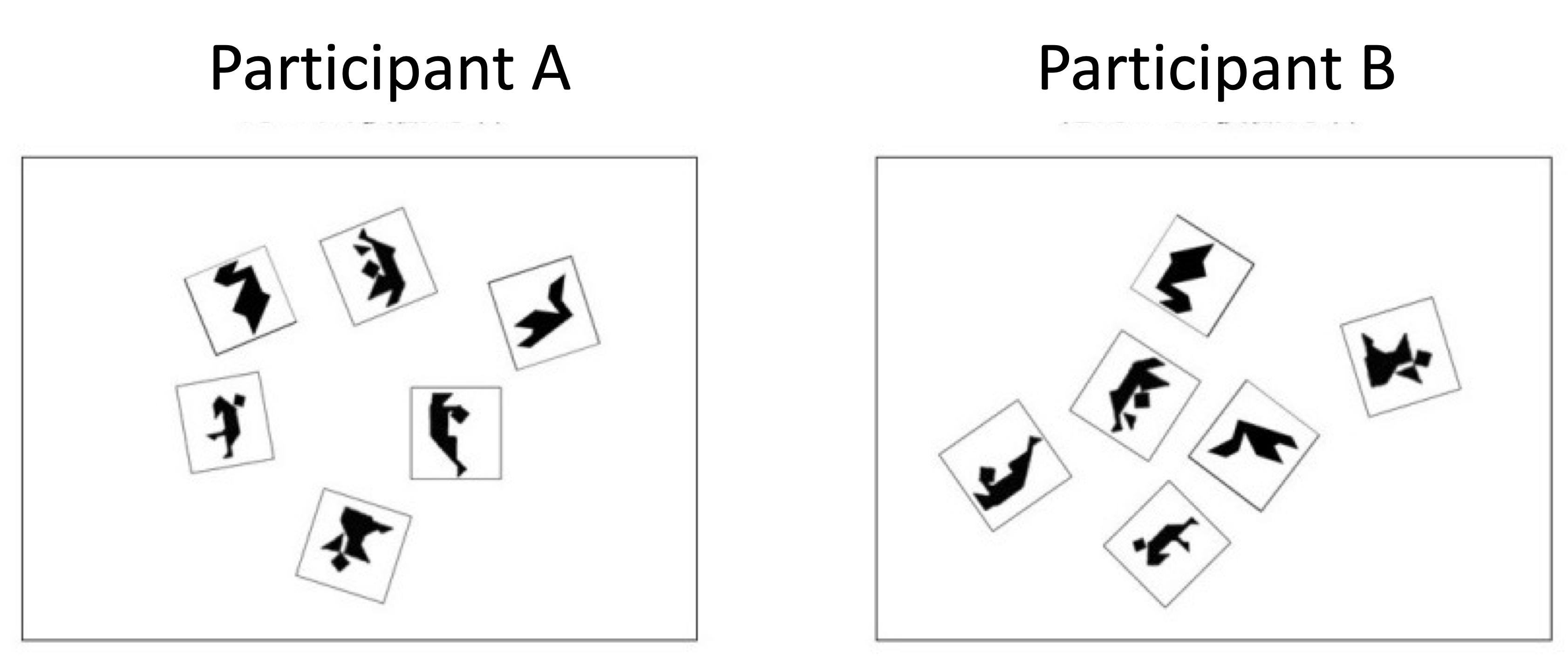}
\caption{An example of placement in the tangram naming task.}
\label{fig:hypothesis}
\end{figure}

\begin{table}[]
    \centering
    \caption{Example dialogue in the tangram naming task}
    \label{tab:my_label}
    \begin{small} 
    \begin{tabular}{c|p{7cm}}
A& And, you know, kicking, like, kicking a ball or something.\\
B& I can't see.\\
A& There is some kind of ball behind the head, and the feet are shaped like the guy is kicking a ball.\\
B& You know, the one with the separate squares?\\
A& Aha, yes, yes, yes.\\
B& Like Hokkaido?\\
A& Hokkaido\\
B& Like a map of Japan\\
A& Oh, no, no, no, no, not that.\\
A& It's kind of a 90-degree kink.\\
B& Ah\\
A& Foot-like, ball-kicking kind of thing.\\
B& Yes, yes, yes, like a little cross-legged thing?\\
A& Oh, yes, and that one with the little square behind it.\\
B& Yeah, I kind of get it.\\
    \end{tabular}
    \end{small}
\end{table}

\section{Model}
\label{sec:contents-format2}

This section presents the concept of a model to simulate the data obtained in TNT. First, we discuss the necessary modules to realize a model in a cognitive architecture. Following this, a detailed model focusing on a specific process in the task is discussed. Finally, preliminary results of simulation are presented.

\begin{figure*}[t]    
\centering
\includegraphics[width=1\textwidth]{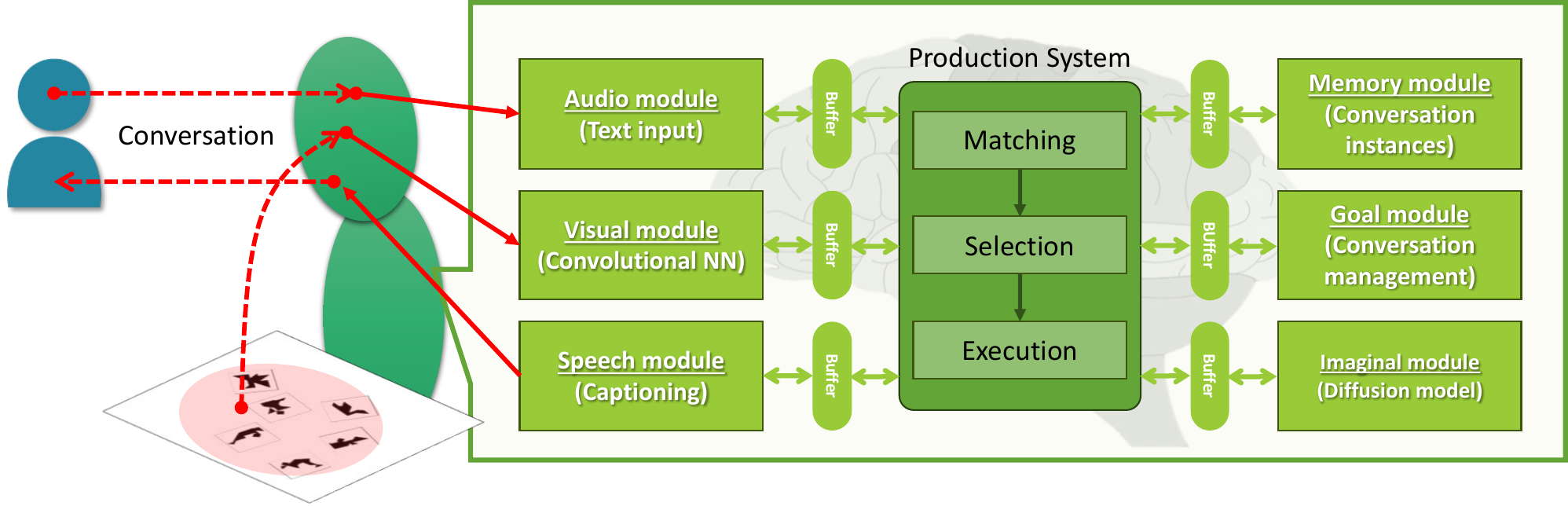}
\caption{Module composition for the tangram naming task.}
\label{fig:hypothesis}
\end{figure*}

\subsection{Module structure}

Figure 2 illustrates the modules assumed to be involved in TNT. In general, cognitive architecture comprises modules related to input/output such as vision, audio, and motor actions; internal process such as holding goal-relevant information and memories; and a central execution unit (working memory or rule engine) that combines the other processes. In Figure 2, the internal modules are placed at the right and the modules related to input/output at the left; they are integrated by a central production system. The role of each module is briefly described below.

\begin{itemize}
    \item {\bf Visual module}: Attend to the tangrams of the external world and recognize them.
    \item {\bf Speech module}: Generate linguistic expressions that distinguish individual tangrams.
    \item {\bf Audio module}: Receive linguistic expressions generated by the other.
    \item {\bf Goal module}: Hold the linguistic expressions assigned to each tangram. It also manages whether or not there is agreement with the other for each expression.
    \item {\bf Imaginal module}: Performs image manipulation to link the expressions to tangrams. The following two types of processing can be assumed for the data of Sudo et al.:
    \begin{itemize}
        \item {\bf Analytical processing}: Decompose a tangram into geometric figures and generate a linguistic representation that allows the partner to identify the tangram.
        \item {\bf Holistic processing}: Generates images based on the language expressions received from the partner.
    \end{itemize}
    \item {\bf Memory module}: Stores examples of past communication. These examples are applied to the current communicative situation to estimate the intention of others. In the study of cognitive architecture, such an intention estimation is known as an instance-based model \cite{gonzalez2003instance} and have been applied to a communication tasks \cite{morita2017implicit,reitter2011groups}. In the context of generative models, examples are considered to be stored as a set of network parameters with a pointer label (ID).
\end{itemize}

\begin{figure}[t]    
\centering
\includegraphics[width=1\columnwidth]{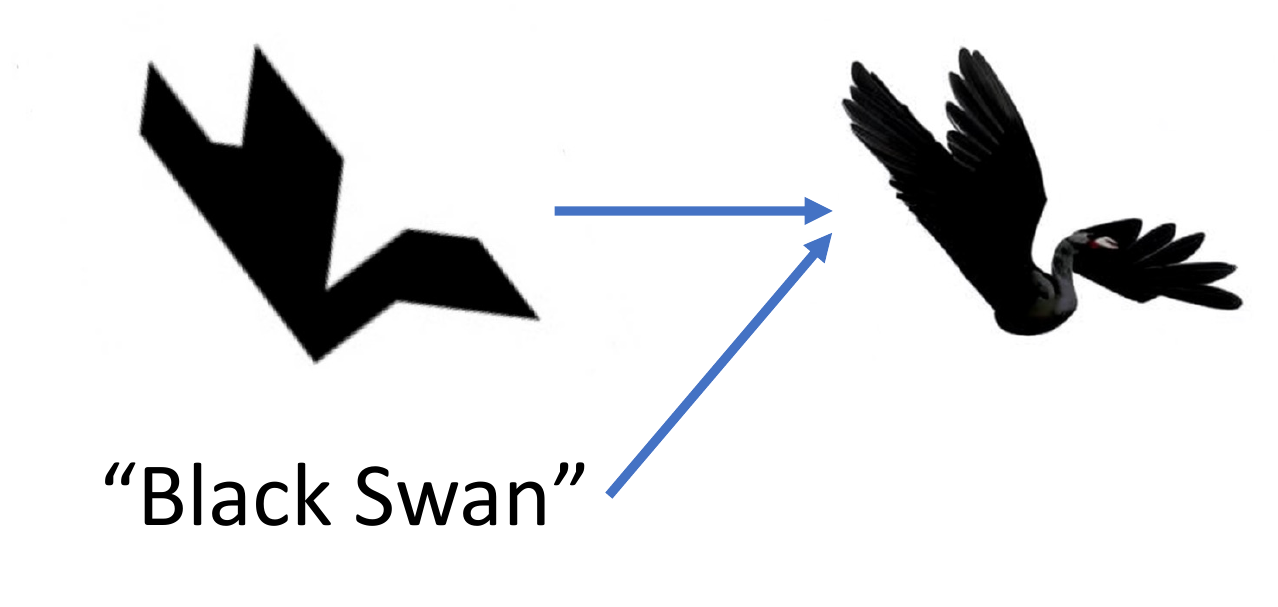}
\caption{Example of images generated by the img2img in Stable Diffusion.}
\label{fig:img2img}
\end{figure}
\begin{figure*}[t]    
\centering
\includegraphics[width=1\textwidth]{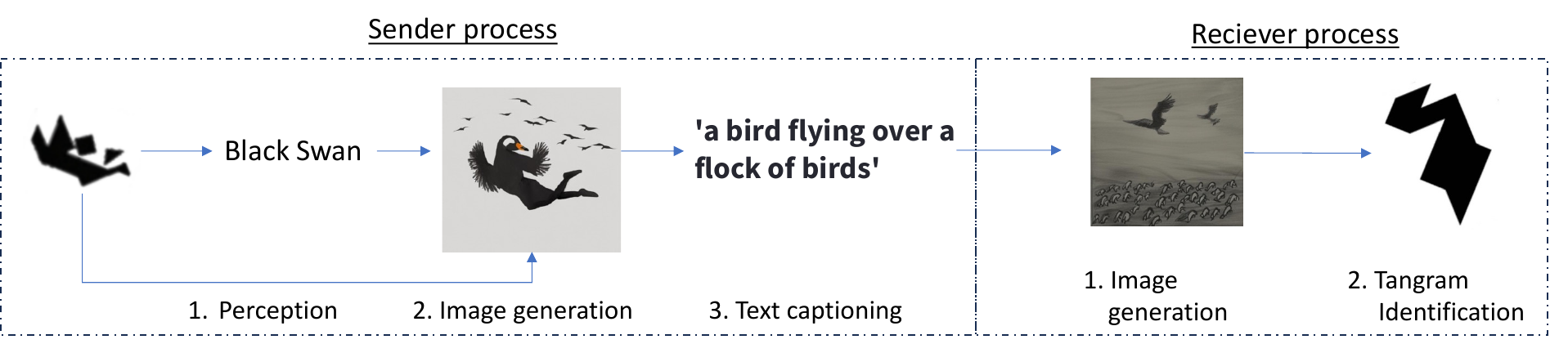}
\caption{Example of holistic process episode.}
\label{fig:img2img}
\end{figure*}

\subsection{One-shot communication}

The overall architecture shown in Figure 2 is large for a single study. To execute such a large project steadily, it is effective to divide the entire model into partial processes. In particular, this paper attempts to detail the sending and receiving of utterances related to the holistic process in the classification of Sudo et al. In the holistic process, the sender generates an image from an observed tangram, linguistic expression from the image, and the receiver reconstructs an image from the received linguistic expression. These processes can be modeled as follows:

\paragraph{Sender process}
\begin{enumerate}
    \item \textbf{Perception (Visual module)}:\\
    Object recognition is performed from the shape of each tangram via the vision module. The vision module is assumed to be implemented by a Convolutional Neural Network (CNN). However, tangram shapes are designed to be polysemous, and there is a possibility that object recognition by ordinary CNNs will encounter difficulties. In particular, for CNN models of general object recognition, the existence of a texture bias (i.e., recognition influenced by the texture of the image surface rather than the holistic shape) has been reported \cite{geirhos2018imagenet}. This bias is considered to have a critical impact on TNT, where no effective cue for classification exists in the texture. Therefore, in this study, we trained a novel CNN to classify 1000 labels from black-and-white images using ImageNet Sketch \cite{wang2019learning} as a dataset that does not introduce texture bias \footnote{A small CNN with six layers (four conv2D and two dense layers) trained with three angled ImageNet sketch dataset (50000 images) reached the validation accuracy of 0.1578 for 1000-class classification. }.
    
    \item \textbf{Image generation (Imaginal module)}:\\
    To generate detailed linguistic expressions beyond mere label from the recognition of tangrams, detailed graphical image is needed. Such a process can be modeled by a generative model called img2img, an image generation component associated with Stable Diffusion \cite{rombach2021highresolution}. Img2img takes a linguistic prompt and an initial image as input and generates a new image. In this study, the input tangram in the previous step is the initial image, and the output label in the previous step is used as the prompt. Figure 3 shows an example of the output of img2img taking a typically observed combination of labels and a tangram in Sudo et al.'s experiment. In this context, the label ``Black Swan'' works as a first impression, which is generated by the visual module and is used to generate the concrete image in the imaginal module.

    \item \textbf{Text captioning (Speech module)}: \\
    As mentioned above, the images constructed by the previous step are considered to be stored in the imaginal module of the architecture shown in Figure 2. Then, the speech module applies image captioning to the images in the imaginal module to obtain detailed linguistic labels. For this process, we use the pre-traind  Vison Encoder Decoder model \cite{kumar2022imagecaptioning} developed based on Vision transformer (ViT) \cite{dosovitskiy2020image} and the GPT-2 \cite{radford2018improving}.
\end{enumerate}

\paragraph{Receiver process}
\begin{enumerate}
    \item \textbf{Image generation (Audio and Imaginal module)}: \\
    The language labels generated by the sender are stored in the receiver’s auditory module. The receiver model generates images from the stored language labels. Stable Diffusion is again used to generate text from the language labels.
    
    \item \textbf{Tangram identification (Visual module)}: \\
    From the image generated in the previous step, the model attempts to identify a tangram image. This process is realized by similarity calculations between the generated image and the observable tangrams. Among the six tangrams, the most matched one is selected in this step. In this study, we calculated a cosine similarity of the output layers of the CNN trained by ImageNet Sketch (same as the sender’s step 1).
\end{enumerate}

Figure 4 shows an example of the abovementioned process. A label ``Black Swan'' is output for the attended tangram by the sender. Then, a detailed image is formed by img2img with the label and the original tangram as input. A caption ``a bird flying over a flock of birds'' is generated for the image. In response to this caption, the receiver produces an image of flying birds that lead to a tangram different from that of the sender.

As described above, the proposed process successfully visualizes the representation transformation sequence occurred in a one-shot communication. Each of these steps is based on existing deep-neural-network models and can be considered to have a certain degree of validity to replicate human performance in specific tasks. However, these processes were subject to varying degrees of errors, and when the processes were combined, the tangram observed by the sender and that identified by the receiver were not the same. In other words, this case visualized an example of miscommunication in TNT.

\subsection{Learning to achieve common grounding}

The process model discussed in the previous section can be improved by repeating the interaction with the same partner. Experiments with human participants have shown that the TNT with the same partner improves communication efficiency \cite{hawkins2020characterizing,sudo-etal-2022-speculative}.


\begin{figure*}[t]
\centering
\includegraphics[width=1\textwidth]{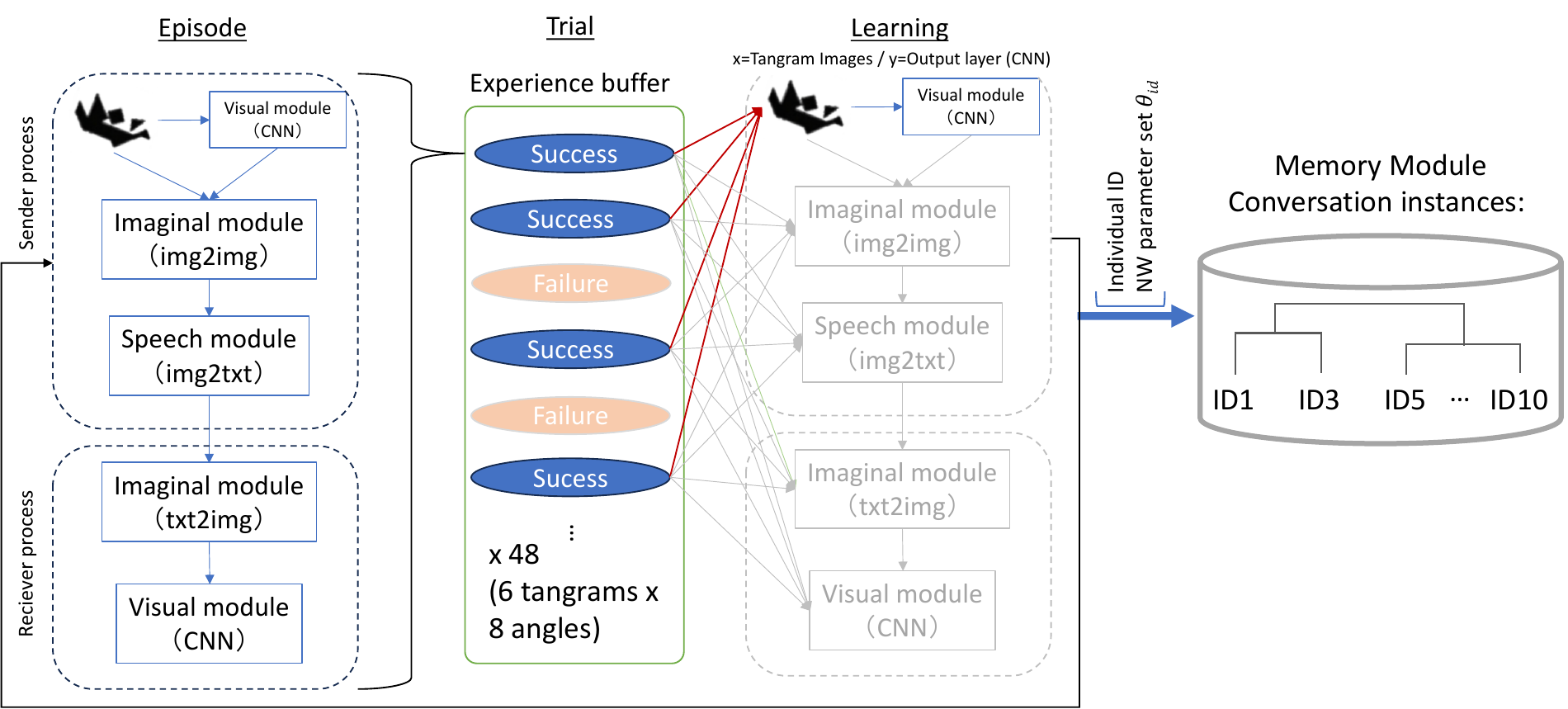}
\caption{Learning process. Each step is assumed to be executed by the module presented in Figure 2.}
\label{fig:learning}
\end{figure*}

The simplified model of building a common ground with the same partner is shown in the left side of Figure 5. The left-most diagram (episode) depicts the process discussed in the previous subsection (the process is vertically transformed from Figure 4). Each episode is applied for each combination of six tangrams and eight angles. For these 48 episodes as one trial, we label the successful cases (where the sender’s perceived tangrams and the receiver’s identified tangrams match) and the unsuccessful cases. Only the successful cases are then used to calibrate the parameters of each component neural network models. For example, in the case of the CNN in the visual module (the highlighted part (red arrows) in the figure), the input tangram image and the output values of the 1000-dimensional vector (output layer) in the successful case are used as test data. Then, the policy gradient algorithm \cite{lapan2018deep}, namely backpropagations minimizing errors between vectors obtained in the successful case and each step of the learning, is applied to adjust the parameters.


The abovementioned process is one involving simple algorithms for deep reinforcement learning, which requires several trials for learning. We assume that humans accumulate such interactions with a specific partner since childhood and hold those experiences as key-value pairs (partnerID-ParameterSet) organized in a hierarchical manner (the right side of Figure 5). In each interaction situation, the accumulated models of past interactions (i.e., interaction schema or cognitive framework) are retrieved to the current situation to enable quick common-ground building.

\section{Preliminary Results}
As already mentioned, the model described by Figures 4 and 5 is only a part of the entire process in TNT. In the actual task, the abovementioned holistic process is combined with other processes such as an analytic process and conversation management process to match the cognitive frameworks of the sender and receiver.

To illustrate the initial state of such a process, we examined the results of the execution of the proposed model. In this study, the sender observed six different tangrams with eight different angles and generated a linguistic expression for each. From those expressions, the receiver generated images and identified the most similar tangram. Figure 6 shows the confusion matrix between the tangrams intended by the sender and those identified by the receiver \footnote{The models used for image generation by the sender and the receiver were v1-5-pruned-emaonly.safetensors, which is the default model in Stable Diffusion. The seed values, which also influence the styles of the images, were set to 1965469825. Due to this setting, no random factors were introduced in this execution.}. The obtained accuracy (percentage of correct responses) was calculated as 0.270, which is higher than the chance rate (8/48 = 0.166) for the six-class classification.


\begin{figure}[t]
\centering
\includegraphics[width=1\columnwidth]{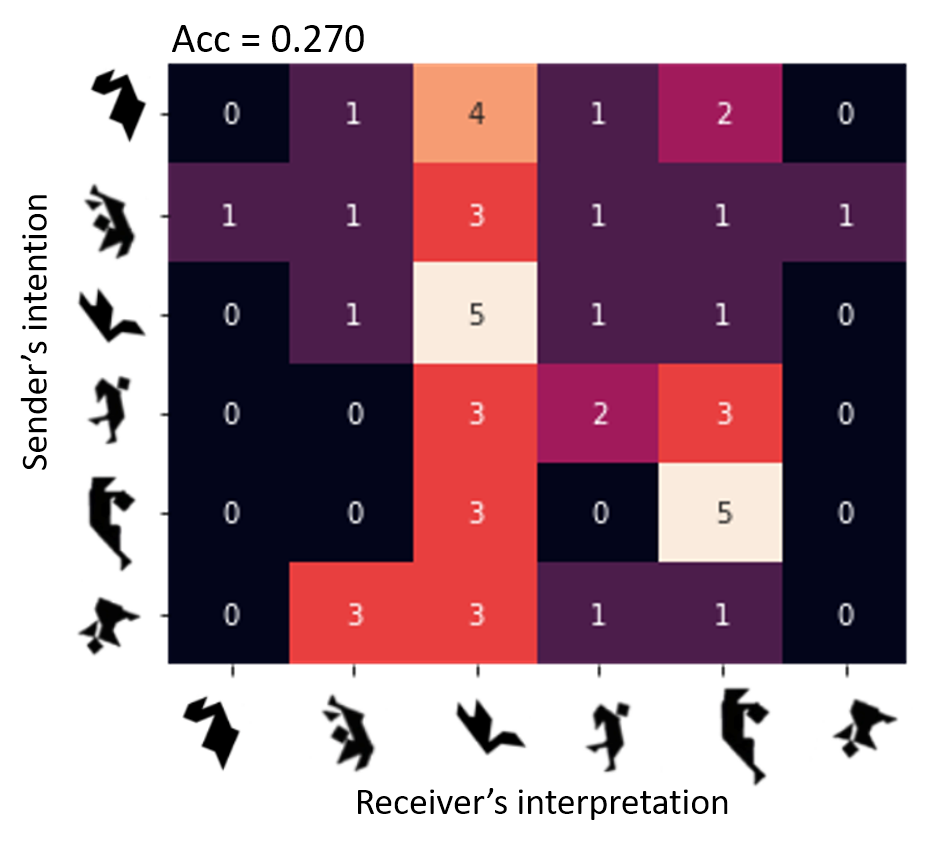}
\caption{Confusion matrix obtained from the initial one-shot communications. }
\label{fig:hypothesis}
\end{figure}

However, even though we achieved improvement beyond the chance level, the model's accuracy remains insufficient for comparison with human common-ground building. In the data collected by Sudo et al., the participating pairs almost perfectly reached agreements on the labels for each tangram. Consequently, we explored whether applying the learning method depicted in the highlighted part of Figure 5, which only tunes parameters in the first CNN, would enhance its accuracy. Figure 7 displays the learning results after 10 repeated trials. The initial trial (0) is set as the value obtained from the aforementioned one-shot communication. The subsequent sequence was obtained through ten independent runs of successive repeated trials. In each learning trial, the batch size and number of epochs were both set to one, and early stopping was applied with a patience level of 10. As depicted in Figure 7, an increase in accuracy percentage can be observed across the majority of runs. A one-sample t-test indicated a statistically significant difference between the mean accuracy obtained (0.288, $n=90$) and the initial level ($t=2.89$, $p<0.01$). In addition, the best cases achieved an accuracy value of 0.395, with one of the corresponding confusion matrices presented in Figure 8. However, this increase remains insufficient to replicate human data in TNT, highlighting the necessity for additional trials or the exploration of learning other network modules.

\begin{figure}[t]
\centering
\includegraphics[width=1\columnwidth]{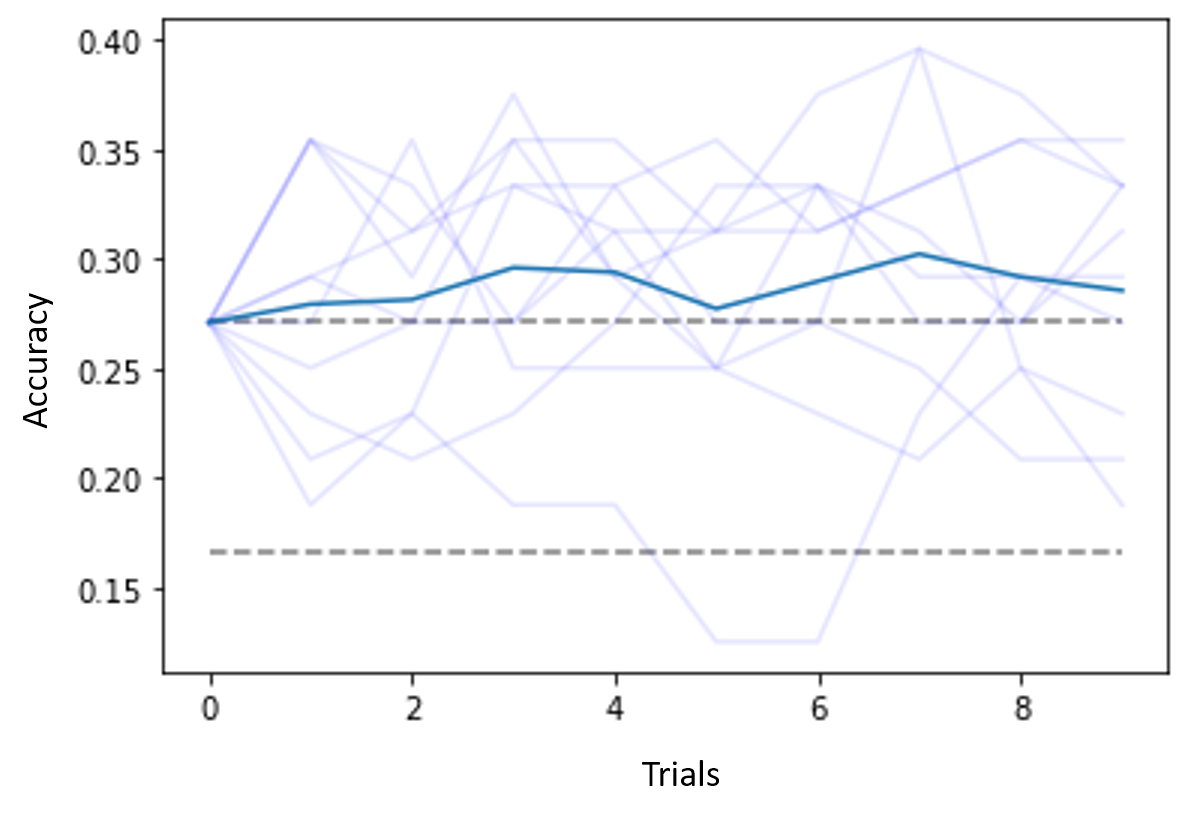}
\caption{Transitions of accuracy. The lower and upper dotted lines indicate the chance and initial levels, respectively. The thin and thick lines
indicate individual and average series, respectively.}
\label{fig:hypothesis}
\end{figure}

\begin{figure}[t]
\centering
\includegraphics[width=1\columnwidth]{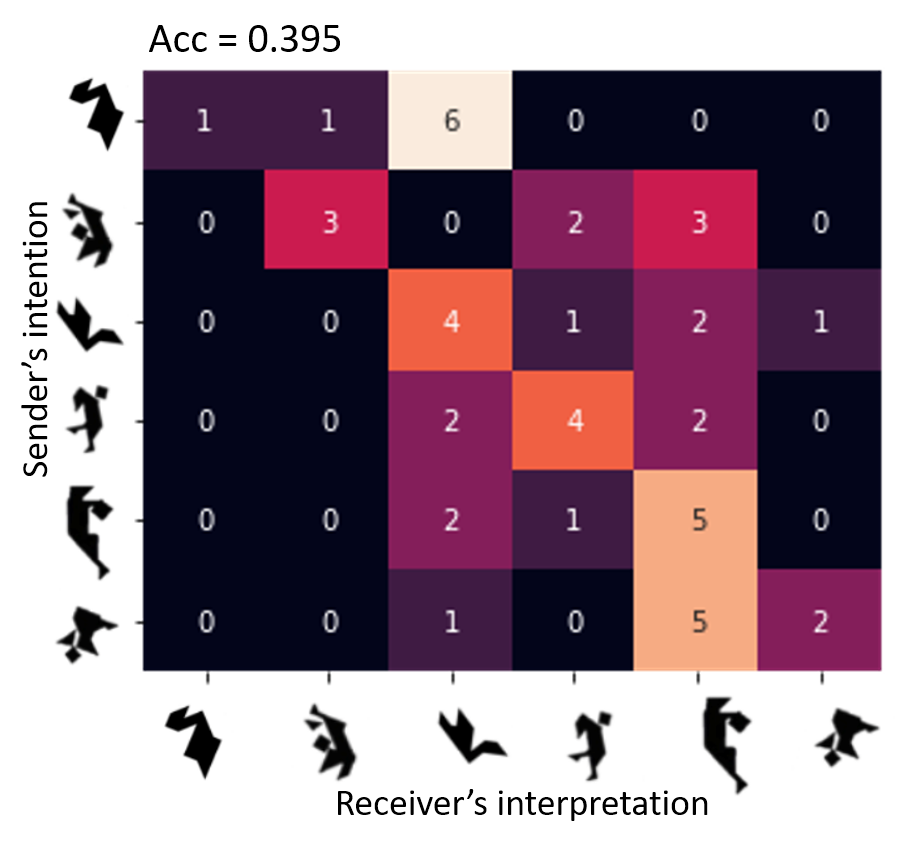}
\caption{One of the best confusion matrices in Figure 7. }
\label{fig:hypothesis}
\end{figure}

\section{Conclusion}


In this paper, we present the concept of a computational model aiming to simulate the cognitive processes behind common-ground building. There have been many preceding studies in the field, reflecting the high interest in the topic. In contrast to preceding studies of common-grounding-building using cognitive architectures \cite{morita2017implicit,reitter2011groups}, this study has advantages using natural language and image generations that are interpretable to humans. In addition, contrary to the previous deep-learning model of TNT \cite{Ji2022:kilogram}, our model is unique in that it is based on the assumption of modules taken from the common cognitive architecture.

By utilizing sub-symbolic knowledge structures embedded in deep-learning models, we prototyped processes from tangram perception to language label generation and from language label receiving to tangram identification. We assumed that this process is mediated by an implicit cognitive framework implemented in network parameters of generative models. By examining the results of the prototype simulation, we expressed the miscommunication between sender and receiver in the TNT and learning through repeated interactions.

The learning results presented in this study are not yet satisfactory, largely because of the lack of computational resources and the difficulty in tuning the network parameters of Stable Diffusion and GPT. 
In addition to addressing this issue, we aim to complete the framework presented herein. We believe that the completion of a model based on the proposed framework will help us to understand the formation of human common ground as well as construct artifacts that share common ground with humans.

\bibliography{nlp2023}
\end{document}